\documentclass{article}

\usepackage{microtype}
\usepackage{graphicx}
\usepackage{booktabs} % for professional tables

\usepackage{hyperref}
\usepackage[ font=footnotesize ]{subfig}

\usepackage[accepted]{icml2025}

\usepackage{amsmath}
\usepackage{amssymb}
\usepackage{mathtools}
\usepackage{amsthm}
\usepackage{algorithmic}
\usepackage{algorithm}
\usepackage{graphicx}
\usepackage{booktabs}
\usepackage{multirow} 
\usepackage{multicol}
\usepackage{subcaption}
\usepackage{longtable}
\usepackage{booktabs}
\usepackage{makecell}
\PassOptionsToPackage{svgnames}{xcolor}
\usepackage{colortbl}
\definecolor{mygray}{gray}{.9}
\usepackage[capitalize,noabbrev]{cleveref}

\theoremstyle{plain}

\theoremstyle{definition}

\theoremstyle{remark}

\usepackage{subfig}

\usepackage{makecell}
\icmltitlerunning{When Dynamic Data Selection Meets Data Augmentation}

\begin{document}

\twocolumn[
\icmltitle{When Dynamic Data Selection Meets Data Augmentation: Achieving Enhanced Training Acceleration}




\begin{icmlauthorlist}
\icmlauthor{Suorong Yang}{nju,ailab}
\icmlauthor{Peng Ye}{ailab,cuhk}
\icmlauthor{Furao Shen*}{nju}
\icmlauthor{Dongzhan Zhou*}{ailab}

\end{icmlauthorlist}

\icmlaffiliation{nju}{National Key Laboratory for Novel Software Technology, Nanjing University}
\icmlaffiliation{ailab}{Shanghai Artificial Intelligence Laboratory}
\icmlaffiliation{cuhk}{The Chinese University of Hong Kong}

\icmlkeywords{Machine Learning, ICML}

\vskip 0.3in
]



\printAffiliationsAndNotice{}  % leave blank if no need to mention equal contribution

\begin{abstract}
Dynamic data selection aims to accelerate training with lossless performance.
However, reducing training data inherently limits data diversity, potentially hindering generalization.
While data augmentation is widely used to enhance diversity, it is typically not optimized in conjunction with selection.
%it is not specifically optimized for augmentation.  
% existing selection methods prioritize representative or challenging samples, which are not specifically optimized for augmentation.  
%As a result, directly combining data selection and augmentation can not effectively leverage the synergies between the two.
As a result, directly combining these techniques fails to fully exploit their synergies.
% Although data augmentation is commonly used to enhance diversity, existing selection methods prioritize representative or challenging samples that are not optimized for augmentation.
% As a result, directly combining selection and augmentation can increase training difficulty and lead to suboptimal performance.
To tackle the challenge, we propose a novel online data training framework that, for the first time, unifies dynamic data selection and augmentation, achieving both training efficiency and enhanced performance.
Our method estimates each sample's joint distribution of local density and multimodal semantic consistency, allowing for the targeted selection of augmentation-suitable samples while suppressing the inclusion of noisy or ambiguous data.
% By estimating each sample's joint distribution of local density and multimodal semantic consistency, our approach selects the most suitable samples for augmentation while avoiding potential noise. 
This enables a more significant reduction in dataset size without sacrificing model generalization.
Experimental results demonstrate that our method outperforms existing state-of-the-art approaches on various benchmark datasets and architectures, e.g., reducing 50\% training costs on ImageNet-1k with lossless performance.
% For instance, on ImageNet-1k, our framework reduces overall training costs by 50\% with lossless performance.
Furthermore, our approach enhances noise resistance and improves model robustness, reinforcing its practical utility in real-world scenarios. 
%Furthermore, leveraging multimodal semantic consistency effectively mitigates the impact of noisy samples, reinforcing the practical utility of our framework for efficient and robust model training in noisy scenarios.
% Furthermore, incorporating multimodal semantic consistency helps reduce the negative impact of potential noisy samples, validating the practical significance of our framework in accelerating model training on real-world datasets.
% The code for reproduction will be made publicly available soon.
\end{abstract}
\begin{figure*}
     \centering
  \includegraphics[width=0.95\textwidth]{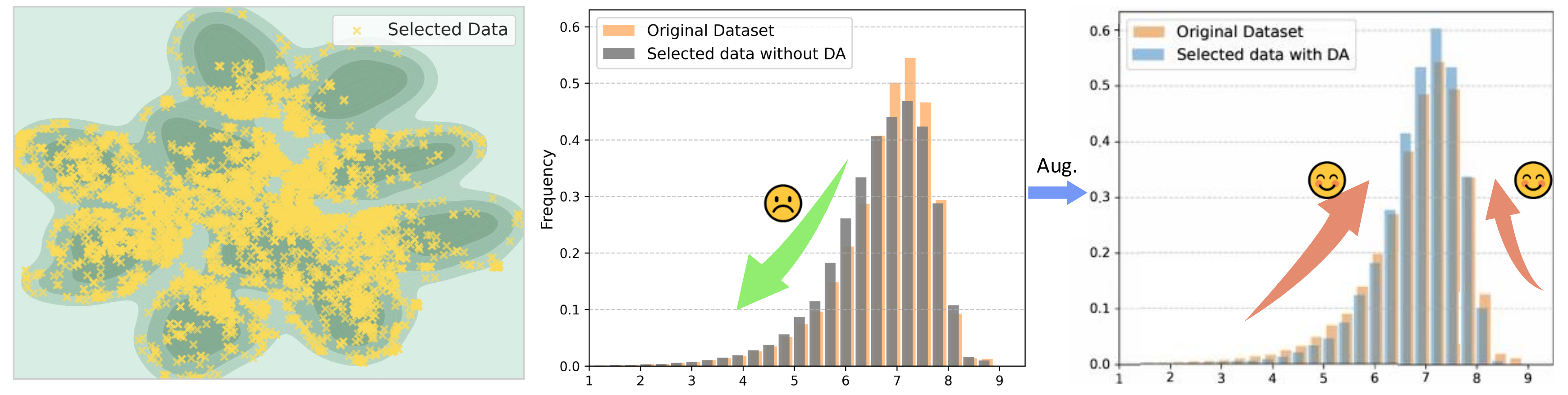}
  \caption{Illustration of the distribution of our selected data points using T-SNE algorithm (\textit{left}) and the density histograms without (\textit{mid}) and with (\textit{right}) augmenting the selected data on CIFAR-10. The selection ratio is 10\%.}
  \label{fig-teaser}
%   \vspace{-6mm}
\end{figure*}
\section{Introduction}
Deep learning has thrived with the growing availability of large-scale datasets. As models become more complex and parameter-extensive, it is necessary to utilize even larger datasets, introducing challenges like reduced training efficiency. Moreover, large-scale datasets often include redundant or noisy samples~\cite{moderate,noise-dataset-1,corrupted_data}, which can compromise training effectiveness.
%Deep learning has greatly benefited from the increasing availability of large-scale datasets. 
%As models grow in complexity and require more parameters, the need for larger datasets to train these models also increases.
%However, the sheer volume of training data presents several challenges, such as training efficiency.
%Additionally, large-scale datasets often contain redundant or noisy samples~\cite{moderate,noise-dataset-1,corrupted_data}, which further hinder training effectiveness.
To address these challenges, various data selection strategies have been proposed to reduce dataset size and enhance data efficiency while maintaining model performance. 
These methods can be broadly categorized into static data selection~\cite{moso,tdds,moderate} and dynamic data selection (or pruning)~\cite{infobatch,dynamic_pruning,dynamic_pruning-2}. 
Static selection identifies a fixed subset of data before training begins, whereas dynamic pruning continuously selects the most influential samples during training.
While these methods can effectively reduce training costs without degrading performance, the reduced training data volume often leads to reduced data diversity.
Consequently, the generalization of deep models is limited, and lossless performance is typically achieved with relatively high selection ratios.

In fact, to address the issue of data diversity, data augmentation is commonly used in model training, which can also improve generalization~\cite{survey3,survey2}.
However, effectively combining data selection and augmentation remains a challenge.
Existing data selection methods typically prioritize representative and challenging samples, which are not specifically designed with augmentation in mind. 
Although augmentation can increase the diversity of selected data and further improve model robustness, applying it to complex samples may introduce ambiguity or noise, potentially increasing training difficulty~\cite{keepaugment,entaugment}.
Therefore, integrating data selection and augmentation in a unified framework presents a promising yet underexplored direction to balance efficiency and generalization.
% Furthermore, leveraging augmentation allows for even lower lossless selection ratios, enabling the removal of more redundant data without sacrificing performance. 
% This synergy not only accelerates training but also aligns with scaling law principles~\cite{scaling}, ensuring that models continue to improve efficiently as they grow in scale with fewer training data.

In this study, we propose a novel framework that integrates dynamic data selection with augmentation, enabling a unified approach to enhance training efficiency and model generalization.
During model training, low-density samples often correspond to underlearned or insufficiently represented data points, such as classification boundaries.
Applying augmentation transformations to these samples reinforces model learning and improves robustness.
However, noisy or outlier samples typically exhibit relatively low density, increasing the risk of introducing noise.
To address this, we introduce a semantic consistency distribution derived from the pre-trained multimodal model CLIP~\cite{clip}.
By prioritizing samples with high sparsity and strong semantic alignment, our approach leverages the joint distribution of density and semantic consistency for effective and robust sample selection.

% Unlike previous approaches that treat data selection and augmentation independently, our method strategically selects samples that are most beneficial for augmentation.
% To identify the sample suitable for data augmentation during training, we estimate the local density of each sample within its neighborhood using an efficient online approximate nearest-neighbor search algorithm, constructing the density distribution of the dataset.
% Since sparse samples often contain ambiguous data, such as noisy or outlier samples, exclusively focusing on these instances inevitably introduces noises.
As illustrated in the left sub-figure of Fig.~\ref{fig-teaser}, the selected data points predominantly cluster around boundary regions among clusters.
Meanwhile, the density histogram in Fig.~\ref{fig-teaser} reveals a more balanced distribution after augmentation, with fewer low- and high-density samples and more data points converging toward the moderate-density regions.
Compared to the distribution without DA, this redistribution highlights our framework's ability to enhance sparse regions, improving model generalization across the entire data distribution.

Experimental results across benchmark datasets and deep architectures demonstrate the effectiveness of our method.
On large-scale datasets such as Tiny-ImageNet~\cite{tiny} and ImageNet-1k~\cite{imagenet}, our method significantly accelerates training while maintaining or even improving generalization.
% Notably, compared to existing baselines, our framework offers two key advantages: i) it achieves lossless performance with lower selection ratios, and ii) it enhances generalization performance when using the same number of training data. 
For instance, on ImageNet-1k, our approach doubles the training efficiency while achieving comparable performance with the entire dataset.
Moreover, our framework exhibits strong cross-architecture and cross-scenario generalization, effectively mitigating the impact of noisy data and enhancing versatility in real-world applications.
On Tiny-ImageNet, our approach outperforms leading baselines by at least 3\% in accuracy under noisy conditions, further demonstrating its reliability.
% These results highlight the effectiveness of our method in improving both training efficiency and model generalization.
% Furthermore, we assess cross-architecture and cross-scenario generalization across ResNet-based~\cite{resnet} and ViT-based~\cite{vit} architectures, as well as in more challenging scenarios such as ImageNet-Hard~\cite{imagenet-hard}, ImageNet-A/O~\cite{imagenet-a}, and ImageNet-R~\cite{imagenet-r}, in addition to noisy data scenarios.
% The results demonstrate that our method achieves superior generalization and robustness while accelerating training.

Our main contributions are summarized as follows: 1) We propose a novel training framework that dynamically integrates data selection and augmentation, significantly accelerating training while maintaining model performance. 2) We introduce a joint distribution based on density and semantic consistency, ensuring effective sample selection and reducing noise and ambiguity. 3) Extensive experiments across diverse datasets and architectures demonstrate superior accuracy and generalization ability, particularly in noisy and challenging scenarios, validating its practical applicability.

\section{Related Work}
\subsection{Data Selection}
The primary goal of data selection is to enhance data-efficient learning, which can be broadly categorized into dataset distillation~\cite{dataset_distillation,dataset_distillation2,dataset_distillation5,dataset_distillation6}, and static or dynamic data selection (or pruning)~\cite{moso,moderate,beyond,infobatch}.
Dataset distillation focuses on synthesizing a small representative dataset that preserves the performance of training on the full dataset.
In contrast, following dynamic data selection without synthesizing new data in this work, we propose a new data training framework that unifies dynamic data selection and augmentation to achieve enhanced model training acceleration.
% \begin{figure}
%   \centering
%   \includegraphics[width=0.99\columnwidth]{./figures/fig1.pdf}
%   %  \hspace{-0.5pt}
%   % \includegraphics[width=0.49\columnwidth]{./figures/density_distribution.pdf}
%   \caption{Illustration of the distribution of our selected data points using T-SNE algorithm (\textit{left}) and the density histograms after augmenting the selected data (\textit{right}) on CIFAR-10. The selection ratio is 10\%.}
%   \label{fig-teaser}
%   \vspace{-6mm}
% \end{figure}

% Following the dynamic data selection, we propose a data training framework that unifies dynamic data selection and augmentation to achieve enhanced model training acceleration.
% In contrast to distillation, which synthesizes an extremely small dataset for training, data selection achieves comparable performance to that obtained on the full dataset while reducing overall training and storage costs.

\textbf{Static data selection} identifies a fixed subset of the training dataset before training begins.
Existing methods can be categorized into selection with importance criteria, dataset distribution-based methods, and optimization-based methods.
\textit{Selection with importance criteria} computes per-sample importance scores and selects the most informative samples.
This includes: 1) the expectation of $\ell_2$-norm error vector and the gradient norm (EL2N and GraNd)~\cite{data_diet}, 2) the change in the optimal empirical risk when a sample is removed~\cite{moso}, 3) the number of forgetting events in the whole training process (Forgetting)~\cite{forgetting}, 4) the impact of including or excluding a sample on the model's classification ability~\cite{score-based-3}.
\textit{Dataset distribution-based methods} select samples based on the geometric distribution of the dataset. 
Herding~\cite{herding} chooses samples based on their distance from the corresponding class centers.
D2~\cite{d2} defines sample difficulty by incorporating the difficulty of its neighboring examples.
The work~\cite{k-center-selection} applies greedy k-center to select the coreset with good data coverage, and CCS~\cite{ccs} balances the sample distribution and importance in selection.
Similarly, Moderate-DS~\cite{moderate} selects samples that are closer to the median score, aiming to balance diversity and representativeness.
\textit{Optimization-based methods} formulate selection as an optimization problem using techniques such as scalable self-supervised pruning metrics~\cite{beyond}, influence function~\cite{dataset_pruning}, bi-level optimization~\cite{glister}, gradient matching~\cite{opt-based-3}, convex optimization~\cite{convex-opt}, facility location function~\cite{crest}, temporal dual-depth scoring~\cite{tdds}, and submodularity~\cite{submodular,cgscore}.

\textbf{Dynamic data selection} identifies informative samples throughout training, allowing the dataset to adapt as the model learns.
% Compared to static selection, dynamic selection offers key advantages: it selects the most influential samples at each training stage, effectively lowering the overall training costs without sacrificing accuracy.
The work~\cite{dynamic_pruning} proposes UCB and $\epsilon$-greedy algorithms to estimate the uncertainty value associated with each training sample, selecting a subset of the data that exhibits the highest levels of uncertainty.
Similarly, the work~\cite{dynamic_pruning-2} also employs both prediction uncertainty and training dynamics to guide the selection process, ensuring that the most informative samples are retained throughout training.
The work~\cite{data-efficient-aug} proposes a data-efficient framework for training neural networks and achieves promising results.
SAS~\cite{data-efficient-contrastive-ssl} improves data efficiency in SSL by proving and selecting the most beneficial data for contrastive training.
Moreover, InfoBatch~\cite{infobatch} proposes a method for unbiased dynamic data selection that accelerates training by pruning less informative samples to retain their relevance in model optimization, which allows for more efficient training without compromising the model’s performance.
% Due to the dynamic nature of sample selection, dynamic data selection can achieve superior performance compared to static selection.

\subsection{Data Augmentation}
Data augmentation (DA) improves the generalization of deep neural networks by increasing the diversity of training samples~\cite{investigating}.
Existing DA methods can be divided into image erasing/mixing-based and automatic augmentation methods~\cite{survey3}.
Image erasing and mixing-based augmentation erase some sub-regions in images or mix random information from two or more images for augmentation to create new samples, respectively.
These methods include Cutout~\cite{cutout}, RandomErasing~\cite{random_erasing}, HaS~\cite{has}, AdvMask~\cite{advmask}, Mixup~\cite{mixup}, GuidedMixup~\cite{guidedmixup}, and GradSalMix~\cite{gradsalmix}, etc.
In addition, based on pre-defined or optimized image transformation policies, automatic DA methods randomly apply one or multiple transformations to each image at each epoch, including AutoAugment~\cite{autoaugment}, Fast-AutoAugment~\cite{fast_autoaugment}, RandAugment~\cite{randaugment}, TrivialAugment~\cite{trivialaugment}, SelectAugment~\cite{selectaugment}, EntAugment~\cite{entaugment}, and MADAug~\cite{madaug}, etc.
Beyond these, generative data augmentation further enriches data by synthesizing new samples using generative models~\cite{generative-da}.
Recent studies also emphasize representation consistency~\cite{atienza2022improving} and address distribution gaps between clean and augmented data~\cite{he2019data}, pointing to new challenges in effective DA design.
% While most automatic DA approaches apply a random combination of transformations to each image, emphasizing the data diversity.
% Most automatic DA approaches apply a random combination of transformations to one image, generating a relatively large degree of transformation.
% Meanwhile, SelectAugment, EntAugment, and MADAug customize the transformation types or magnitudes for training samples during training, which brings additional computation costs to model training.
\begin{figure*}
    \centering
    \includegraphics[width=1.0\linewidth]{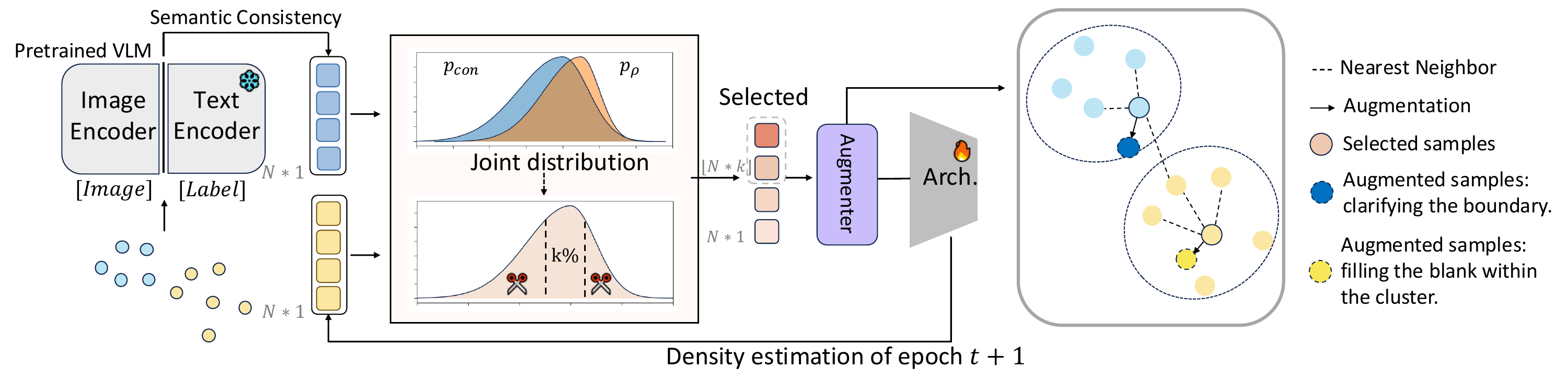}
    \caption{The framework of our proposed data training method: The core idea of our framework is to construct a joint distribution that integrates both the density and semantic consistency distributions, enabling the prioritization of low-density, semantically consistent samples. After augmentation, augmented sparse samples in intra-cluster regions help to fill the underrepresented spaces, while samples located around the decision boundaries between clusters differentiate the classification decision more clearly, thus improving generalization.}
    \label{fig:framework}
\end{figure*}
% Among these methods, TrivialAugment offers a computationally efficient alternative by applying only a single, lightweight transformation per image.
% This approach provides two key advantages: i). It introduces negligible computation overhead to the online training process, making it highly efficient.
% ii). Since each image undergoes only one transformation with light magnitude, the augmented samples remain close to their original feature space.
% This is well-suited for our dynamic data selection framework, which targets samples in underrepresented regions.
% By ensuring that augmented samples remain within their original data distribution, TrivialAugment complements data selection without disrupting model training. 
% Therefore, we adopt TrivialAugment as our augmentation strategy.

\section{The Proposed Method}
\subsection{Overview of the Proposed Method}
As shown in Fig.~\ref{fig:framework}, we propose a novel data training framework that integrates dynamic data selection with augmentation to enhance both training efficiency and generalization.
Our framework employs two complementary distributions: 1). a density distribution, dynamically estimated by a task-specific model (e.g., ResNet), which identifies underrepresented samples for augmentation and training, and (2) a semantic consistency distribution, computed using a frozen pre-trained multimodal model (CLIP), which quantifies the alignment between an image and its corresponding textual label.
Low-density regions highlight underrepresented samples but may also include noisy or ambiguous instances. To address this issue, the semantic consistency distribution acts as a strong complement, filtering samples with weak semantic alignment.
By combining both distributions, we construct a joint distribution that captures the relationship between sample informativeness and semantic correctness.
Consequently, samples with higher joint distribution scores are prioritized to be selected and augmented for training.

\subsection{How to Identify Samples for Augmentation}
% Existing data selection methods aim to identify a subset of representative samples, denoted as $\hat{D} \subseteq D$.
% While training on $\hat{D}$ reduces training costs, it also compromises data diversity, potentially limiting model generalization. 
% To mitigate this, data augmentation is commonly used to introduce diversity.
% However, selecting the right samples for augmentation remains a challenge.

Given a dataset $D$ that follows an underlying distribution $P(D)$, the optimization objective of our dynamic data selection module at time $t$ is to select a subset $\hat{D}_t$, containing at most $k$ samples.
The goal is to minimize the expected loss over the distribution $P(D)$, with the following optimization: $\hat{D}_t = \underset{\hat{D}_t \subseteq D,\left|\hat{D}_t\right| \leq k}{\arg \min } \mathbb{E}_{z \sim P(D)}\left[L\left(z, \hat{\theta}_{\mathcal{A(}\hat{D}_t) }\right)\right]$,
where $z$ represents a test sample, $L$ is the loss function, $\hat{\theta}_{\hat{D}_t}$ is the empirical risk minimizer on $\hat{D}_t$, and $\mathcal{A}$ represents the augmentation operations applied to the selected subset.
% By incorporating augmentation into the selected data, we increase the diversity of the data, further enhancing the model’s generalization.
% A critical challenge, therefore, is to identify which samples are most suitable for augmentation.

Our method prioritizes low-density samples because such regions in the feature space often correspond to underlearned or insufficiently represented data points.
Focusing on these samples and applying augmentation operations helps the model capture distinctive features.
By augmenting these sparse samples, we compensate for their underrepresentation, improving the model’s ability to generalize across diverse data regions.
To efficiently determine the density of data points during online training, we exploit an online approximate nearest neighbor search architecture (HNSW)~\cite{hnsw} to query the nearest $k$ neighbors of each sample $x$, denoted as $NN(x)$.
The density of a sample is then estimated as the mean of the $\ell_2$ distance between $x$ and its neighbors:
\begin{equation}\label{eq:density_cal}
    \rho_{x_i} = \frac{1}{k} \sum_{j\in NN(x_i)} || x_i - x_{j} ||,
\end{equation}
where higher values indicate low-density samples. 
The density scores are then normalized using a Min-Max scaling to obtain $p_\rho(x)$.
For augmentation, we employ slight augmentation, which generates neighboring samples within low-density areas while preserving local structure.

% However, aggressive augmentation may inadvertently push samples into high-density areas, contradicting the goal of selection.

Nevertheless, low-density regions may also contain challenging, outlier, or noisy samples.
Continuously selecting these samples for training can substantially complicate training, especially on real-world datasets that inevitably include noise. 
To address this issue and improve the practical effectiveness of our method, we introduce a multimodal semantic consistency constraint that simultaneously refines the selection of low-density samples.
This ensures that augmentation is applied to meaningful data, improving both robustness and efficiency in the training process.

\begin{table*}[]
    \centering
    \caption{The accuracy (\%) comparison to state-of-the-art baselines. All methods are trained with ResNet-18 on CIFAR-10/100 and ResNet-50 on Tiny-ImageNet. Note that some results could not be computed due to the unavailability of open-source code and parameter settings, making it impossible to reproduce. Random* means randomly selecting samples in each epoch.\label{tab:comparison_experiment}}
    % \vspace{-1mm}
	% \renewcommand\arraystretch{.9}
	\resizebox{0.75\textwidth}{!}{
    \begin{tabular}{c|ccc|ccc|ccc}
    \bottomrule[1.5pt]
    Dataset &  \multicolumn{3}{c|}{\cellcolor{mygray}CIFAR-10}& \multicolumn{3}{c|}{\cellcolor{mygray}CIFAR-100} & \multicolumn{3}{c}{\cellcolor{mygray}Tiny-ImageNet} \\ \hline
    Whole Dataset &\multicolumn{3}{c|}{95.6}&\multicolumn{3}{c|}{78.2} &\multicolumn{3}{c}{45.0} \\ \hline
    Selection Ratio (\%)& 30& 50&70& 30& 50&70& 30& 50&70 \\ \hline
    
    Random &90.2&92.3&93.9&69.7&72.1&73.8&29.8&37.2&42.2 \\ 
    Herding~\cite{herding} &80.1 &88.0&92.2 &69.6&71.8&73.1&29.4&31.6&39.8 \\
    EL2N~\cite{data_diet} &91.6&95.0&95.2 &69.5&72.1&77.2 &26.6&37.1&44.0 \\
    GraNd~\cite{data_diet} &91.2&94.6&95.3 &68.8&71.4&74.6 &29.7&36.3&43.2 \\
    Glister~\cite{glister} &90.9&94.0&95.2 &70.4&73.2&76.6 &30.1&39.5&43.9 \\
    Forgetting~\cite{forgetting} &91.7&94.1&94.7 &69.9&73.1&75.3 &28.7&33.0&41.2 \\
    Moderate-DS~\cite{moderate} &91.5&94.1&95.2&70.2&73.4&77.3 &30.6&38.2&42.8 \\
    Self-sup. prototypes~\cite{beyond} &91.0&94.0&95.2&70.0&71.7&76.8 &27.7&37.9&43.4 \\
    MoSo~\cite{moso} &91.1&94.2&95.3 &70.9&73.6&77.5 &31.2&38.5&43.4 \\
    DP~\cite{dataset_pruning}&90.8&93.8&94.9 &-&73.1&77.2 &-&-&- \\
    Random* &93.0&94.5&94.8 &74.4&75.3&77.3 &41.5&42.8&43.1 \\
    UCB~\cite{dynamic_pruning} &93.9&94.7&95.3 &-&75.3&77.3 &-&-&-    \\
    $\epsilon$-Greedy~\cite{dynamic_pruning} &94.1&94.9&95.2 &-&74.8&76.4&-&-&- \\
    InfoBatch~\cite{infobatch} &94.7&95.1&95.6 &76.5&78.1&78.2 &42.2&43.2&43.8 \\ 
    % CVPR (\textbf{Pruning}) &\textbf{94.9}&95.1&95.9&77.4&78.7&78.9&44.7&44.9&45.1 \\
    % ICLR (\textbf{Selection})&90.8&&95.0 &72.5&&77.0 &31.7&&46.0 \\
    \hline
    Ours &\textbf{94.9}&\textbf{95.5}&\textbf{96.0}&\textbf{77.6}&\textbf{78.9}&\textbf{79.5}&\textbf{44.9}&\textbf{47.0}&\textbf{49.4} \\
    % Ours &\textbf{94.9} &\textbf{95.1} &\textbf{95.9} &\textbf{77.4}&\textbf{78.7}& \textbf{78.9}&\textbf{44.7}&\textbf{44.9}&\textbf{45.1} \\
    \bottomrule[1.5pt]
    \end{tabular}}
    \vspace{-1mm}
\end{table*}
\subsection{Multimodal Consistency Estimation for Robust Data Selection}
Noisy data - arising from incorrect labels, corrupted images, or outlier samples - reflects a fundamental mismatch between the semantic content of $x$ and its corresponding label $y$. 
%Therefore, the data cleaner assesses the joint distribution $p(x,y)$, which captures the plausibility of an image-label pair.
To detect and filter such inconsistencies, the data cleaner evaluates the joint distribution $p(x,y)$, which captures the plausibility of an image-label pair.
%Given the multimodal nature of $x$ and $y$ (i.e., images and text), 
Given the inherent correlation and multimodal nature of $x$ and $y$, we introduce multimodal consistency supervision as an additional criterion for assessing sample reliability.
This complements the density-based selector by filtering out samples that exhibit low cross-modal consistency.

To implement this, we leverage a pre-trained CLIP model to embed images and text into a shared multimodal space, enabling semantic alignment assessment.
However, CLIP's zero-shot generalization is limited to domain-specific datasets, making it necessary to adapt the embeddings to the target domain.
Instead of performing computationally expensive fine-tuning, we incorporate lightweight adapters of MLP for both the image and text encoders~\cite{adapters,mllm}.
The lightweight architecture of adapters ensures efficient adaptation while preserving CLIP's pretrained knowledge.

To measure the cross-modal consistency, we compute the cosine similarity of the encoded image and text features:
\begin{equation}
con(x_i) = \ell_{cos}(E_I(x_i), E_T(y_i)),
\end{equation}
where $E_I$ and $E_T$ are visual and textual encoders, respectively.
The consistency scores are normalized via Min-Max scaling to approximate the consistency distribution $p_{con}(x)$, where higher values indicate stronger semantic alignment. Since the image-label alignment is derived from a pretrained vision-language model and remains independent of the training process, we precompute the consistency distribution beforehand.
This enables direct use during sample selection, eliminating additional computational overhead in online training.
%The image-label alignment, derived from the pretrained vision-language model, is independent of the training process
%Therefore, to ensure high efficiency during online training, we precompute the consistency distribution before training, allowing it to be directly used during sample selection without incurring additional overhead.
 
\subsection{Augmenter}
To integrate structural sparsity and semantic consistency, we define a joint distribution that combines both density and consistency distributions:
\begin{equation}\label{eq:joint-dis}
    p_{sel}(x_i) = p_\rho(x_i) * p_{con}(x_i).
\end{equation}
Here, $p_\rho$ evolves dynamically with model training, allowing sample selection to adapt to the current model training state. 
During online training, samples with higher joint distribution scores are prioritized for augmentation and training, ensuring that both underrepresented and semantically meaningful samples are utilized effectively.

We employ TrivialAugment as our augmenter, which is widely used and offers a computationally efficient augmentation strategy.
During augmentation, only a single lightweight transformation per image is applied.
This brings two key advantages: i). It introduces negligible computation overhead to the online training process, making it highly efficient. ii). Since each image undergoes only one transformation with a slight magnitude, the augmented samples remain within their original local feature space. This is well-suited for the objective of our dynamic data selection framework: filling intra-cluster gaps and enhancing decision boundaries within clusters.
Thus, the consistency of selected samples is preserved while the data diversity in sparse regions is enhanced.
Consequently, training using these augmented samples improves model performance.
In addition, we provide the augmentation operation list used in Table~\ref{tab:augmentation_ops} in the Appendix, which includes image transformations such as translation, rotation, equalization, etc.

\paragraph{Complexity analysis.}
The computational costs of our framework, when integrated into online training, are primarily associated with density estimation.
Specifically, both querying and updating within the HNSW graph operates with a complexity of $\mathcal{O}(\log(n))$, where $n$ is the total number of data points.
Let $T$ denote the total number of training epochs; then, the total cost is $\mathcal{O}(T * \log(n))$.
Since $T \ll n$, the overall computational complexity remains $\mathcal{O}(\log(n))$, making our method scalable for large datasets.
Furthermore, the data augmentation, as a standard pipeline in model training, introduces negligible overhead.

% to efficiently determine the local sparsity of training samples, we exploit an online approximate nearest neighbor search architecture (HNSW) to query the nearest $k$.
% The average distance between the $i$-th query sample $\boldsymbol{x}_i$ and its corresponding neighbor samples $NN(\boldsymbol{x}_i)$ is calculated as the sparsity score $s_i$, i.e., $s_i = \frac{1}{k}\sum_{\boldsymbol{x'}_j \in NN(\boldsymbol{x}_i)}\ell_2(\boldsymbol{x}_i, \boldsymbol{x'}_j)$.
% Therefore, a sample with higher sparsity scores indicates its higher local sparsity in feature space.
% It is more beneficial to apply augmentation methods to these training samples.

% However, sparse samples in feature space typically exist among noisy or difficult data. Applying strong augmentation could introduce ambiguous data. Following~\cite{}, the multimodal foundation models are used to filter and suppress such samples.
% Based on the multimodal information, we aim to select both representative and sparse samples for augmentation.
% The representativeness can be formulated as the conditional distribution $p(y|\boldsymbol{x})=p(y,\boldsymbol{x})/p(y)$. In classification tasks, $p(y)$ is independent of the specific training samples

\section{Experiment}
\subsection{Experiment Setup}
\paragraph{Datasets and network architectures.} In line with previous works~\cite{moso,moderate,infobatch}, we evaluate the effectiveness of our proposed method using widely adopted benchmark datasets, including CIFAR-10/100~\cite{cifar100}, Tiny-ImageNet~\cite{tiny}, and ImageNet-1k~\cite{imagenet}.
In addition, we evaluate the robustness of our method in noisy datasets.
To further assess the generalization ability of our method, we extend the evaluation to more challenging datasets, such as ImageNet-A/O~\cite{imagenet-a}, ImageNet-Hard~\cite{imagenet-hard}, and ImageNet-R~\cite{imagenet-r}.
Additionally, we evaluate the generalization of our method across different deep architectures.
Specifically, we conduct experiments using both ResNet-based, such as ResNet-18/50, and ViT-based models, such as ViT-B/L and Swin-Transformer, to demonstrate the robustness and scalability of our approach across diverse models.

\begin{table*}[]
       \centering
    \caption{Results on ImageNet-1k with a 60\% selection ratio using ResNet-50 on an 8-A100 server. Note that due to the high computational costs and device memory costs~\cite{moderate}, Glister and CG-Score are not reported. Some results are from~\cite{infobatch}. Time is the wall clock time; Overall (n*h)  is the total GPU hour, where $n$ is the node number.\label{tab:imagenet-1k} }
    % \vspace{-1mm}
	\resizebox{0.95\textwidth}{!}{
    \begin{tabular}{c|ccccccccccc|cc}
    \toprule[1.5pt]
    Method &Herding&EL2N&GraNd&Forgetting&SSP&Moderate&MoSo&UCB&Infobatch&Glister&CG-Score&Ours&Whole Dataset \\ \hline
    Acc. (\%) &71.1&72.3&71.0&72.5&70.0&73.1&74.0&75.8&76.5&-&-&\textbf{76.9}&76.4 \\ \hline
    Time (h) &10.5&10.5&10.5&10.5&10.5&10.5&10.5&10.5&10.5&10.5&10.5&10.5&17.5 \\
    Overhead (h) &$>$17.5&$>$17.5&$>$17.5&$>$17.5&$>$24.0&$>$17.5&$>$17.5&0.03&0.0028&-&-& 0.53&0.0 \\
    Overall (n*h) & $>$224.0 &$>$224.0&$>$224.0&$>$224.0&$>$276.0&$>$224.0&$>$224.0&\textbf{84.0}&\textbf{84.0}&-&-&88.2 &140.0 \\
    \bottomrule[1.5pt]
    \end{tabular}}
    \vspace{-3mm}
\end{table*}

\begin{figure*}[h]
    \begin{minipage}{0.31\textwidth}
        \captionof{table}{Experimental results on Tiny-ImageNet with noisy and corrupted data using ResNet-50. The noisy ratio is 20\%. }
		\label{tab:noisy-dataset}
    \centering
    \setlength{\tabcolsep}{2.5pt}
        \resizebox{.95\textwidth}{!}{\begin{tabular}{c|cc|cc}
          \bottomrule[1.1pt]
        \multirow{2}{*}{\makecell{Method / \\Selection Ratio (\%)}} 
        & \multicolumn{2}{c|}{\cellcolor{mygray} Noisy}
        & \multicolumn{2}{c}{\cellcolor{mygray} Corrupted} \\ \cline{2-5}
        & 20 & 30 & 20 & 30 \\ \hline
        Random & 17.8 &23.9 &20.0&25.9 \\  
        Herding & 19.0&24.2&35.0&30.6 \\
        Moderate-DS &19.6&25.0&23.3&29.1 \\
        EL2N &13.9 &18.6 &18.6&24.4 \\
        GraNd &18.3&23.7&20.0&26.7 \\
        Forgetting &13.2&21.8&18.5&25.5 \\
        Self-sup. prototypes &15.1&21.0&20.2&26.9 \\
        CG-Score &8.4&15.3&16.4&24.4 \\
        Glister & 21.6&25.5&21.2&22.0\\
        MoSo &7.4&11.3&23.1& 28.8\\
        Random* &33.8&36.5&35.1&36.9 \\
        InfoBatch &34.9&37.1&35.1&38.1 \\ \hline
        Ours &\textbf{35.9}&\textbf{39.6}&\textbf{39.1}&\textbf{42.0} \\
         \bottomrule[1.1pt]
        \end{tabular}}
    \end{minipage}
    \hspace{3mm}
    \begin{minipage}{0.31\textwidth}
        \centering
    \captionof{table}{Experimental results on Tiny-ImageNet with data augmentation. The selection ratios are 30\%, 50\%, and 70\%. \label{tab:effect-augmentation}}
    \resizebox{.95\textwidth}{!}{
    \begin{tabular}{c|ccc}
    \bottomrule[1.5pt]
    Whole Dataset &\multicolumn{3}{c}{52.0} \\ \hline
Selection Ratio & 30\% &50\% &70\% \\ \hline
  Random &29.8&37.2 &42.2 \\
  Herding &31.6&39.2&45.6 \\
  EL2N & 32.0&40.1&45.9 \\
  GraNd &32.2&40.5&46.2 \\
  Glister &33.1&42.2&46.5 \\
  Forgetting &27.2&36.2&44.2 \\
  Moderate-DS &33.8&41.5&46.6 \\
  Self-sup. proto. &33.4&41.1&46.6 \\
  MoSo &32.6 &41.5 &45.9 \\
  Random* &42.1&43.9&45.2 \\ 
  InfoBatch &43.2&45.9&48.3 \\ \hline
  % CVPR & $\sim$ 44.9  &46.49&49.18 \\
  % ICLR &33.96&&47.76 \\
  Ours &\textbf{44.9}&\textbf{47.0}&\textbf{49.4} \\
      \bottomrule[1.5pt]
    \end{tabular}}
    \end{minipage}
    \hspace{3mm}
    \begin{minipage}{0.31\textwidth}
        \centering
    \includegraphics[width=0.8\linewidth]{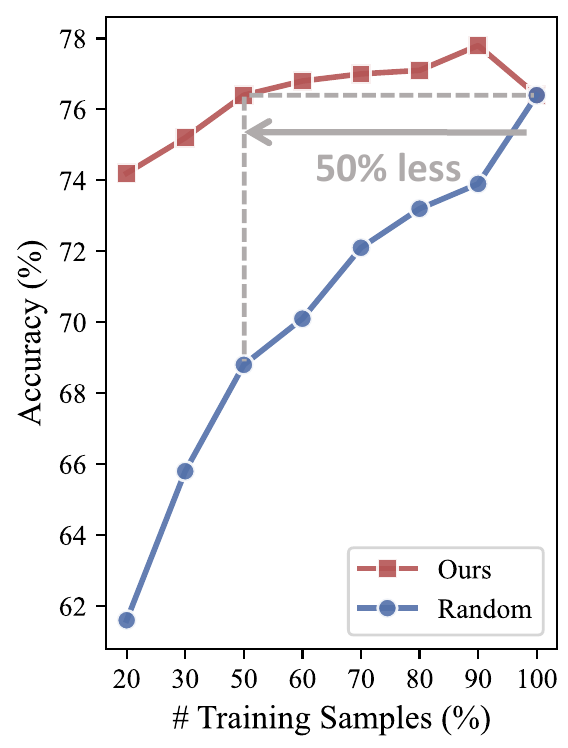}
    \vspace{-5mm}
    \caption{The performance on ImageNet-1k across various selection ratios with a 4-A100-GPU server.}
    \label{fig:imagenet-1k-sel-ratio}
    \end{minipage}
     \vspace{-.5cm}
\end{figure*}

\paragraph{Comparison with state-of-the-arts.}
We compare with our method both static and dynamic data selection methods, including 1) Random, 2) EL2N~\cite{data_diet}, 3) GraNd~\cite{data_diet}, 4) Herding~\cite{herding}, 5) Forgetting~\cite{forgetting}, 6) Moderate-DS~\cite{moderate}, 7) Self-sup. prototypes~\cite{beyond}, 8) MoSo~\cite{moso}, 9) DP~\cite{dataset_pruning}, 10) UCB~\cite{dynamic_pruning}, 11) $\epsilon$-Greedy~\cite{dynamic_pruning}, 12) Glister~\cite{glister}, and 13) InfoBatch~\cite{infobatch}.

\paragraph{Implementation details.}
To ensure consistency with prior work~\cite{infobatch, moderate}, we follow similar experimental settings.
Specifically, we use the OneCycle scheduler with the SGD/LARS optimizer for model training, a momentum of 0.9, a weight decay of 5e-4, and cosine annealing.
We employ TrivialAugment~\cite{trivialaugment} in our framework.
For fairness, we adopt the annealing and re-scaling techniques introduced in~\cite{infobatch}, which standardize the dynamic dataset pruning process across all methods compared.
Moreover, we use InfoNCE loss to fine-tune adapters for 15 epochs on all datasets.
Since InfoBatch uses soft pruning with a dynamic number of selected samples, we report its performance using the same number of forward passes as in our method.

\subsection{Performance Comparison}
As shown in Table~\ref{tab:comparison_experiment}, we evaluate the performance of our method by training ResNet-18 on CIFAR10/100 and ResNet-50 on Tiny-ImageNet across different selection ratios.
Our method achieves comparable performance to models trained on the full dataset, even when only 50\% of the data is used on CIFAR-10/100 and 30\% on Tiny-ImageNet.
In contrast, existing methods typically achieve lossless data selection with relatively higher selection ratios on these datasets, such as over 60\% on CIFAR-10/100 and over 70\% on Tiny-ImageNet.

Notably, our approach outperforms the other methods at the same selection ratios.
On Tiny-ImageNet, a large-scale dataset, our method yields an average performance improvement of at least 2.7\% while maintaining the same training costs.
As the training data volume increases, this performance gap becomes even more pronounced, further highlighting the efficiency and effectiveness of our framework.
\begin{table*}[]
    \centering
    \caption{Generalization of models trained with our method on ImageNet-Hard, ImageNet-A, ImageNet-R, and ImageNet-O. We report AUPR (\%) for ImageNet-O and accuracy (\%) on others. All models are ResNet-50.}
    \label{tab:imagenet-variant}
    \resizebox{0.77\textwidth}{!}{
    \begin{tabular}{cccccccc|c}
    \toprule[1.5pt]
  Selection Ratio(\%)&20 &30& 50 & 60 & 70   & 80 &  90 & Whole Dataset \\ \hline
      ImageNet-A &1.9\scriptsize{$\downarrow$1.2}&2.1\scriptsize{$\downarrow$1.0}&2.9\scriptsize{$\downarrow$0.2}&3.1\scriptsize{$\uparrow$0.0}&3.4\scriptsize{$\uparrow$0.3}&3.4\scriptsize{$\uparrow$0.3}&\textbf{3.5\scriptsize{$\uparrow$0.4}}&3.1 \\
      ImageNet-R&37.2\scriptsize{$\uparrow$1.0}&38.5\scriptsize{$\uparrow$2.3} &39.3\scriptsize{$\uparrow$3.1}&39.8\scriptsize{$\uparrow$3.6}&39.9\scriptsize{$\uparrow$3.7}&40.6\scriptsize{$\uparrow$4.4}&\textbf{41.0\scriptsize{$\uparrow$4.8}}&36.2 \\
      ImageNet-O&15.4\scriptsize{$\uparrow$2.2}&15.8\scriptsize{$\uparrow$2.6} &16.1\scriptsize{$\uparrow$2.3}&16.3\scriptsize{$\uparrow$2.5}&16.3\scriptsize{$\uparrow$2.5}&16.4\scriptsize{$\uparrow$2.6}&\textbf{16.5\scriptsize{$\uparrow$2.7}}&13.2 \\
      ImageNet-Hard&14.2\scriptsize{$\downarrow$0.5}&15.3\scriptsize{$\uparrow$0.6}&15.9\scriptsize{$\uparrow$1.2}&16.5\scriptsize{$\uparrow$1.8}&16.7\scriptsize{$\uparrow$2.0}&17.2\scriptsize{$\uparrow$2.5}&\textbf{17.5\scriptsize{$\uparrow$2.8}} &14.7 \\
      \bottomrule[1.5pt]
    \end{tabular}}
    \vspace{-3mm}
    \vspace{-0.2cm}
\end{table*}

\subsection{ImageNet-1k Results}
Table~\ref{tab:imagenet-1k} presents the evaluation results of our method on the ImageNet-1k dataset with a 60\% selection ratio.
Our approach outperforms the full dataset by achieving a nearly 40\% training cost reduction, resulting in a reduction of up to 56 hours in training overhead with a 0.5\% accuracy improvement.
Meanwhile, since most static data selection methods require training surrogate models to determine the sample's influence throughout model training, the computation overheads are relatively much higher than ours.
Thus, the results highlight that our method outperforms static data selection methods in both performance and computational efficiency.
It also surpasses dynamic pruning methods in terms of final accuracy with comparable efficiency.
These findings underscore the generality and competitiveness of our approach to large-scale datasets.

Further analysis of the performance across different selection ratios on ImageNet-1k is shown in Fig.~\ref{fig:imagenet-1k-sel-ratio}.
The results show that our method achieves lossless performance with only 50\% of the training data. 
When using 20\% of the data, performance drops by about 2\% while nearly 80\% of the training overhead is eliminated.
Compared to random selection, which suffers a significant accuracy drop as the selection ratios decrease, our method maintains robust performance even with reduced data.
Similarly, most existing baseline methods typically require at least 60\% of the training data to achieve similar lossless performance. This demonstrates that our framework further lowers the data requirement for maintaining full performance.
%Consequently, our proposed framework further lowers the lossless data selection ratio.

%  \begin{table}[]
%  \caption{Experimental results on Tiny-ImageNet with noisy and corrupted data using ResNet-50. The noisy ratio is 20\%. }
% 		\label{tab:noisy-dataset}
%     \centering
%     \setlength{\tabcolsep}{2.5pt}
%         \resizebox{.32\textwidth}{!}{\begin{tabular}{c|cc|cc}
%           \bottomrule[1.1pt]
%         \multirow{2}{*}{\makecell{Method / \\Selection Ratio (\%)}} 
%         & \multicolumn{2}{c|}{\cellcolor{mygray} Noisy}
%         & \multicolumn{2}{c}{\cellcolor{mygray} Corrupted} \\ \cline{2-5}
%         & 20 & 30 & 20 & 30 \\ \hline
%         Random & 17.8 &23.9 &20.0&25.9 \\  
%         Herding & 19.0&24.2&35.0&30.6 \\
%         Moderate-DS &19.6&25.0&23.3&29.1 \\
%         EL2N &13.9 &18.6 &18.6&24.4 \\
%         GraNd &18.3&23.7&20.0&26.7 \\
%         Forgetting &13.2&21.8&18.5&25.5 \\
%         Self-sup. prototypes &15.1&21.0&20.2&26.9 \\
%         CG-Score &8.4&15.3&16.4&24.4 \\
%         Glister & 21.6&25.5&21.2&22.0\\
%         MoSo &7.4&11.3&23.1& 28.8\\
%         Random* &33.8&36.5&35.1&36.9 \\
%         InfoBatch &34.9&37.1&35.1&38.1 \\ \hline
%         Ours &\textbf{35.9}&\textbf{39.6}&\textbf{39.1}&\textbf{42.0} \\
%          \bottomrule[1.1pt]
%         \end{tabular}}
%         \vspace{-3mm}
% \end{table}

\subsection{Robustness to Noisy Scenarios}
In real-world scenarios, training data is often polluted by corrupted and mislabeled images~\cite{noisy_labels,corrupted_data}, which can significantly degrade model performance.
Specifically, we simulate mislabeled data (noisy) by flipping a portion of the labels to incorrect ones using symmetric label noise.
Meanwhile, we introduce five types of realistic distortions to simulate corrupted data, namely Gaussian noise, random occlusion, resolution variations, fog, and motion blur. 
The examples of corrupted data are shown in the Appendix.
To evaluate the practical relevance of our data training framework in such noisy environments, we assess the robustness of our method compared to existing state-of-the-art methods.

As shown in Table~\ref{tab:noisy-dataset}, our approach consistently outperforms the compared methods, demonstrating superior robustness against both mislabeled and corrupted data.
Specifically, our method achieves a 4\% improvement over competing methods on corrupted datasets, even with a 20\% noise ratio on Tiny-ImageNet.
Our framework excels in these scenarios due to its ability to combine low-density sample selection with multimodal semantic alignment. 
While noisy data is typically sparse and low-density, our method's robust integration of multimodal semantics offers a powerful mechanism for mitigating noise and highlighting meaningful patterns in the data.
This approach allows us to maintain high robustness without sacrificing data efficiency.

By prioritizing sparse, low-density samples and leveraging the corrective power of multimodal alignment, our method provides a reliable and efficient solution for robust deep learning in practical, noisy data environments, demonstrating its practical significance.

\subsection{Effect of Data Augmentation on Model Performance}
To further assess the effectiveness of our method, we compare its performance against several baseline methods using TrivialAugment for data augmentation, as shown in Table~\ref{tab:effect-augmentation}.
Our method consistently outperforms the other approaches across various selection ratios.

While data augmentation enhances the performance of all methods, our approach consistently achieves superior results at different selection ratios.
This indicates that our method is not simply a straightforward combination of data augmentation and data selection.
Instead, it effectively identifies the most beneficial samples for augmentation, leading to significant performance improvements.
By selectively amplifying the impact of data augmentation, our method optimizes model performance, demonstrating its ability to leverage augmentation more effectively than other approaches.

\subsection{Generalization on Hard Benchmarks}
To evaluate the generalization capabilities of our proposed framework, we conduct experiments on challenging benchmark datasets, including ImageNet-Hard~\cite{imagenet-hard}, ImageNet-R~\cite{imagenet-r}, and ImageNet-A/O~\cite{imagenet-a}. 
Specifically, we pre-train ResNet-50 models using data selected through our method across various selection ratios and then test their performance on these challenging benchmark datasets.
Following standard evaluation settings, we report the area under the precision-recall curve (AUPR) for ImageNet-O and classification accuracy for the other datasets.

As shown in Table~\ref{tab:imagenet-variant}, our method maintains or even enhances generalization performance on these challenging datasets, despite using fewer training samples. 
The results demonstrate that reducing the dataset size with our framework does not compromise generalization ability. 
Meanwhile, as the selection ratios increase, our method achieves superior generalization compared to training on the full dataset.
\begin{table}[]
    \centering
    \vspace{-2mm}
    \caption{Experiment results on more advanced architectures, including ViT-B, ViT-L, and Swin-T on ImageNet-1k with a 4-A100 GPU server. Overhead represents GPU hours (h), and $S_r$ refers to the selection ratio.}
    \label{tab:imagenet-vit}
    \resizebox{0.42\textwidth}{!}{
    \begin{tabular}{c|ccccc|c}
    \toprule[1.5pt]
     $S_r$(\%)&50 & 60 & 70 & 80 & 90& Full Dataset \\ \hline
     % R-18 &&&&&&& \\
     ViT-B &82.6&82.9&83.2&83.2&\textbf{83.3}&82.5 \\
     ViT-L &85.2&85.3&85.6&\textbf{85.7}&\textbf{85.7}&84.6 \\
     Swin-T &84.1&84.1&84.2&84.2&\textbf{84.3}& 84.2\\
      \bottomrule[1.5pt]
    \end{tabular}}
    \vspace{-5mm}
\end{table}

\subsection{Generalization on Different Architectures}
To evaluate the scalability of our proposed method, we conduct experiments on advanced architectures, including ViT-B, ViT-L~\cite{vit}, and Swin-T~\cite{swin-t}.
Specifically, we train these architectures using our framework across various selection ratios.

As shown in Table~\ref{tab:imagenet-vit}, our framework is architecture-agnostic, achieving robust generalization across these different models, even with reduced selection ratios.
Notably, the lossless performance can be achieved with only 50\% of the training data. The results underscore that our method generalizes on ResNet-based and Transformer-based architectures, all with reduced training costs.

\begin{figure}
    \centering
    \includegraphics[width=0.8\linewidth]{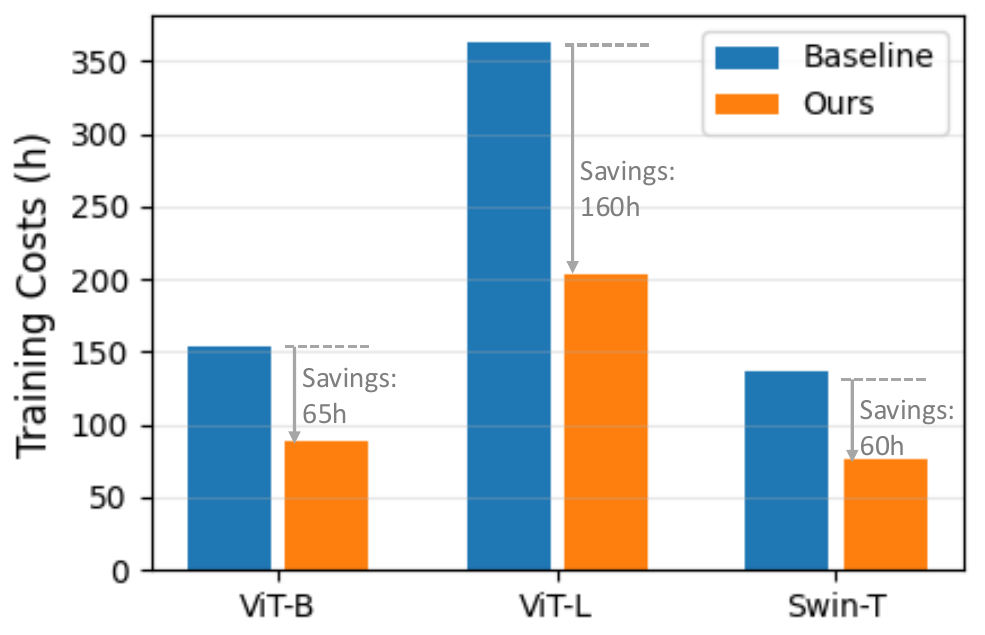}
    \vspace{-2mm}
    \caption{The overall cost savings achieved by our method on ViT-based architectures with lossless performance. The experiment is conducted on ImageNet-1k with a 4-A100-GPU server.}
    \label{fig:cost-reduction}
    \vspace{-.2cm}
\end{figure}

Additionally, in Fig.~\ref{fig:cost-reduction}, we present the practical training costs on these architectures and the lossless cost savings achieved by our framework.
It can be seen that our proposed framework can significantly save hundreds of hours on large-scale architecture training.
\begin{figure}[]
	\centering
    \includegraphics[width=0.9\linewidth]{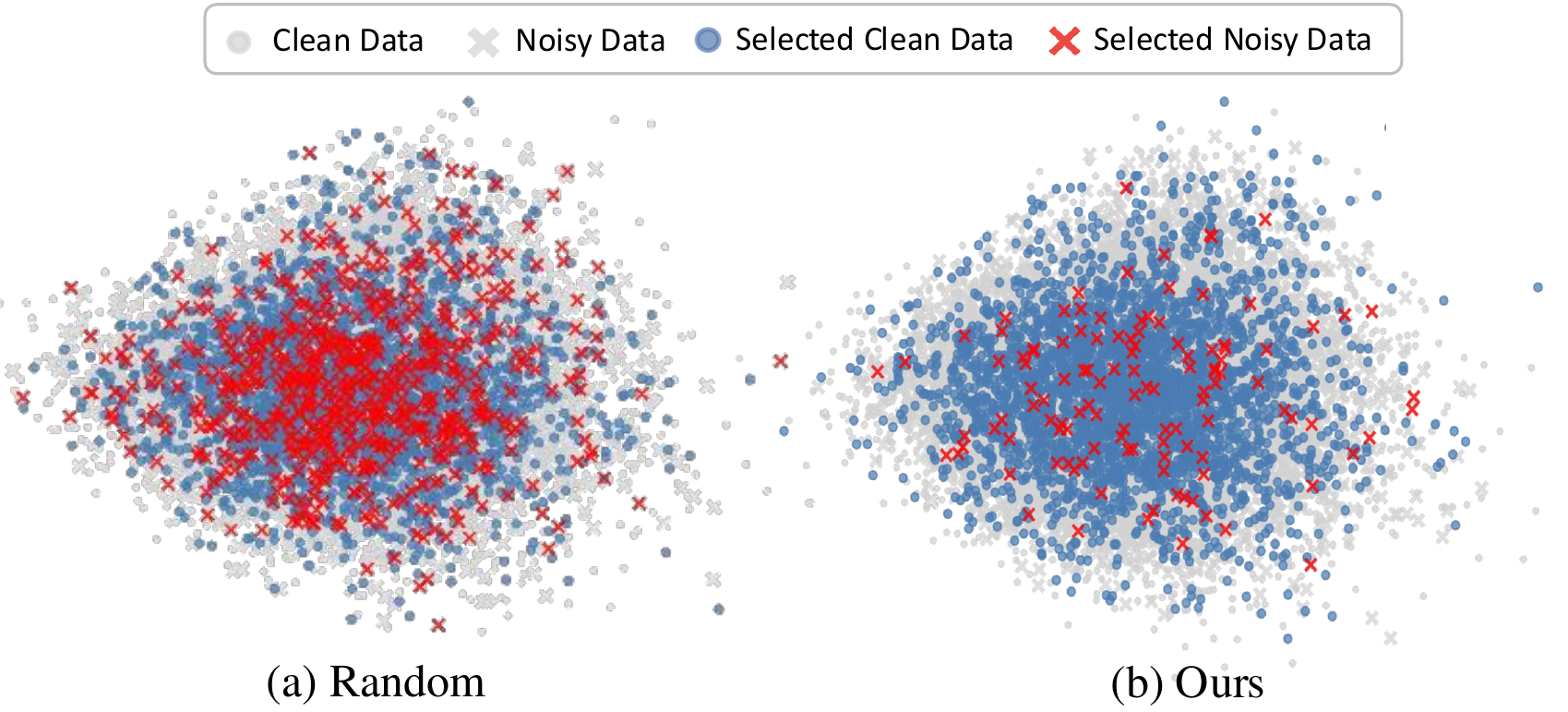}
	\caption{Visualization of the selection results on noisy Tiny-ImageNet with a 20\% noise ratio. The selection ratio is 20\%.}
    \vspace{-7mm}
	\label{fig:vis-selection}
\end{figure}
% \begin{figure}[]
% 	\centering
%     \vspace{-0.1cm}
%         \subfloat[Random.]{\label{fig5-1}
%         \includegraphics[width=0.5\columnwidth]{./figures/random_selected_results.pdf}
%             }  
%             \hspace{-5mm}
%        \subfloat[Ours.]{\label{fig5-2}  
%            \includegraphics[width=0.5\columnwidth]{./figures/ours_selected_results.pdf}
%                } 
% 	\caption{Visualization of the selection results on noisy Tiny-ImageNet with a 20\% noise ratio. The selection ratio is 20\%.}
%     \vspace{-7mm}
% 	\label{fig:vis-selection}
% \end{figure}
\subsection{Visualization of the Selection Robustness}
In Fig.~\ref{fig-teaser}, we illustrate our selection results on clean datasets, showing that the selected data points mainly cluster around boundary regions among clusters.
To better understand our selection effectiveness, in Fig.~\ref{fig:vis-selection}, we further illustrate our selection results on the noisy Tiny-ImageNet dataset with a 20\% noise ratio.
It can be seen that compared to the baseline, our method can effectively filter out noisy samples: the number of selected noisy points is minimized.

\subsection{Further Analysis of the Overheads}
Although our method introduces negligible overheads into online training, our framework incorporates adapter fine-tuning and feature embedding via CLIP models and corresponding adapters before model training begins.
As shown in Table~\ref{tab:overhead}, we analyze these pre-computation overheads of the adapter fine-tuning and feature embedding via CLIP models.
It can be observed that these one-time overheads before model training are negligible compared to standard target model training. 
Once computed, no further computation is required during online training across selection ratios.

\subsection{Ablation Study}
\paragraph{Effect of different modules in our framework.}
In Table~\ref{tab:effect-supervision}, we systemically evaluate the effectiveness of different components within our proposed framework on Tiny-ImageNet using ResNet-50 across various selection ratios.

When only the density distribution $p_\rho$ is used, performance is lower, as low-density samples often include sparse and outlier data, which can introduce ambiguity into the training process.
However, when both the density distribution $p_\rho$ and consistency distribution $p_{con}$ are combined, performance improves, demonstrating that incorporating semantic consistency helps mitigate the negative effects of density-based selection.
Further performance gains are achieved by including the augmentation module, which boosts accuracy by a significant margin.  This shows that augmentation plays a crucial role in improving performance, especially when the selected data points under $p_\rho$ and $p_{con}$ are more suited for augmentation, enhancing the model’s generalization.
The results indicate that removing any module from our framework leads to a substantial drop in performance.

\begin{table}[]
    \centering
    \caption{Overheads of fine-tuning and feature embedding before model training on large-scale datasets with a 1-V100 GPU server.}
    \label{tab:overhead}
    \resizebox{0.42\textwidth}{!}{
    \begin{tabular}{c|ccc}
    \toprule[1.5pt]
     Dataset & Fine-tuning & Embedding & Overall training \\ \hline
     Tiny-ImageNet &0.39h&0.03h&21.0h \\
     ImageNet-1k &1.25h&0.17h&84.0h \\
      \bottomrule[1.5pt]
    \end{tabular}}
    \vspace{-4mm}
\end{table}

\begin{table}[]
    \centering
    \caption{Effect of density distribution, consistency distribution, and augmenter on Tiny-ImageNet using ResNet-50. We report test accuracy (\%). The selection ratios (\%) are 30\%, 50\%, and 70\%. \label{tab:effect-supervision}}
    \resizebox{0.3\textwidth}{!}{
    \begin{tabular}{ccc|ccc}
    \bottomrule[1.5pt]
   $p_\rho$ & $p_{con}$ & aug. & 30\% &50\% &70\% \\ \hline
   \checkmark& &&39.0 & 40.7 & 42.5 \\
   &\checkmark&&42.0 & 45.6 & 45.8 \\
   &&\checkmark&41.6 & 45.9 & 48.3 \\
   & \checkmark & \checkmark & 42.5 & 46.3 & 49.3 \\
   \checkmark & \checkmark& & 41.5 & 43.1 & 44.3 \\
   \checkmark && \checkmark & 41.1 & 45.1 & 48.5 \\
   \checkmark &\checkmark& \checkmark &\textbf{43.5} & \textbf{47.5} & \textbf{50.2} \\
      \bottomrule[1.5pt]
    \end{tabular}}
    \vspace{-3mm}
\end{table}

% 单独的D和A都不能保证样本选择的合理性，尤其是仅有D，low D往往会引入稀疏的离群样本，给训练带来歧义；
% 选择class-representative samples + aug能取得较好的结果，但是性能上限有局限性；
% 在这个基础上，加入了alignment的约束，能显著提升效果。

% \paragraph{Cross-augmentations Generalization.}
% In this section, to demonstrate the cross-augmentation generalization of our proposed method, we employ various widely-used data augmentation methods in our proposed framework, including Cutout~\cite{cutout}, AutoAugment~\cite{autoaugment}, RandAugment~\cite{randaugment}, TrivialAugment~\cite{}, and EntAugment~\cite{entaugment}.

% \begin{table}[]
%     \centering
%     \caption{Cross-augmentation generalization analysis by integrating different SOTA DA baselines into our framework on Tiny-ImageNet using ResNet-50. \label{tab:effect-DA-methods}}
%     \resizebox{0.39\textwidth}{!}{
%     \begin{tabular}{c|ccc}
%     \bottomrule[1.5pt]
% Selection Ratio & 30\% &50\% &70\% \\ \hline
% w/o Aug. &39.7&42.7&44.3 \\
% % RE &38.4&41.2&43.5 \\
%  FAA & 40.5$\uparrow 0.8$ & 44.1$\uparrow 1.4$ & 46.9$\uparrow 2.6$ \\
%  Cutout & 39.8$\uparrow 0.1$ & 43.2$\uparrow 0.5$ & 44.3$\uparrow 0.0$ \\
%  DADA & 41.5$\uparrow 1.8$ & 45.5$\uparrow 2.8$ & 48.7$\uparrow 4.4$ \\
%  AA & 40.3$\uparrow 0.6$ & 44.7$\uparrow 2.0$ & 47.7$\uparrow 3.4$ \\
%  RA & 40.4$\uparrow 0.7$ & 45.4$\uparrow 2.7$ & 47.2$\uparrow 2.9$ \\
%  TA &44.9$\uparrow 5.2$ &47.0$\uparrow 4.3$&49.4$\uparrow 5.1$ \\
%       \bottomrule[1.5pt]
%     \end{tabular}}
% \end{table}
\section{Conclusion}
This paper proposes a novel data training framework that unifies dynamic data selection and data augmentation for more enhanced model training acceleration.
Unlike existing selection methods, our proposed approach identifies samples suitable for data augmentation.
By combining this with augmentation, our framework can improve model generalization with reduced training costs.
As a result, we can achieve lossless training acceleration with fewer data and enhanced generalization using the same volume of data.
Extensive experiments demonstrate the effectiveness and efficiency of our method, especially in terms of generalization across large-scale datasets and more challenging scenarios.

% \section*{Impact Statement}
% This paper introduces a novel data training framework that unifies dynamic data selection and data augmentation, aimed at improving the training efficiency and generalization of deep learning models.

% By reducing data requirements and accelerating training, our framework can significantly lower the computational resources and time needed to train large-scale models, which is a key consideration for academic research and industry applications. 
% This could lead to more accessible machine learning technologies, particularly in resource-constrained environments, by reducing dependency on large datasets and extensive computational infrastructure.

% Ethically, our approach promotes efficiency without compromising model performance, which could have positive implications for environmental sustainability by reducing the carbon footprint associated with training large models.

% In the future, as deep learning models, such as LLM, continue to scale, our work could contribute to more sustainable practices in AI development while maintaining high standards of accuracy and fairness.

\nocite{langley00}

\bibliography{example_paper}
\bibliographystyle{icml2025}

%%%%%%%%%%%%%%%%%%%%%%%%%%%%%%%%%%%%%%%%%%%%%%%%%%%%%%%%%%%%%%%%%%%%%%%%%%%%%%%
%%%%%%%%%%%%%%%%%%%%%%%%%%%%%%%%%%%%%%%%%%%%%%%%%%%%%%%%%%%%%%%%%%%%%%%%%%%%%%%
% APPENDIX
%%%%%%%%%%%%%%%%%%%%%%%%%%%%%%%%%%%%%%%%%%%%%%%%%%%%%%%%%%%%%%%%%%%%%%%%%%%%%%%
%%%%%%%%%%%%%%%%%%%%%%%%%%%%%%%%%%%%%%%%%%%%%%%%%%%%%%%%%%%%%%%%%%%%%%%%%%%%%%%
% \newpage
% \input{appendix}

% You can have as much text here as you want. The main body must be at most $8$ pages long.
% For the final version, one more page can be added.
% If you want, you can use an appendix like this one.  

% The $\mathtt{\backslash onecolumn}$ command above can be kept in place if you prefer a one-column appendix, or can be removed if you prefer a two-column appendix.  Apart from this possible change, the style (font size, spacing, margins, page numbering, etc.) should be kept the same as the main body.
%%%%%%%%%%%%%%%%%%%%%%%%%%%%%%%%%%%%%%%%%%%%%%%%%%%%%%%%%%%%%%%%%%%%%%%%%%%%%%%
%%%%%%%%%%%%%%%%%%%%%%%%%%%%%%%%%%%%%%%%%%%%%%%%%%%%%%%%%%%%%%%%%%%%%%%%%%%%%%%

\end{document}